\documentclass[letterpaper, 10 pt, conference]{ieeeconf}  

\IEEEoverridecommandlockouts 
\usepackage{amsfonts}       
\usepackage{amsmath}
\usepackage{bm}
\usepackage{booktabs}       
\usepackage{tabularx}
\usepackage{times}
\usepackage{epsfig}
\usepackage{graphicx}
\usepackage{multirow}
\usepackage{makecell}
\usepackage{nicefrac}       
\usepackage{xcolor}         
\usepackage{algorithm}
\usepackage{algorithmic}
\usepackage{url}
\usepackage{etoolbox,lipsum}
\usepackage{indentfirst}
\usepackage[dvipsnames]{xcolor}
\usepackage{xspace}
\usepackage{bbm}
\usepackage{amssymb}
\usepackage{microtype}
\usepackage{pifont}
\usepackage{cuted}
\usepackage{cite} 
\usepackage{stfloats}
\usepackage{hyperref}

\newcommand{\ourmethod}{\textup{DIAL-GS}\xspace}

\title{\LARGE \bf
\ourmethod: \underline{D}ynamic \underline{I}nstance \underline{A}ware Reconstruction for \underline{L}abel-free Street Scenes with 4D Gaussian Splatting}

\author{
    Chenpeng Su$^{1*}$,
    Wenhua Wu$^{1*}$,
    Chensheng Peng$^{2}$,
    Tianchen Deng$^{1}$,
    Zhe Liu$^{1}$,
    Hesheng Wang$^{1\dagger}$
}

\begin{document}

\maketitle
\thispagestyle{empty}
\pagestyle{empty}

{
\makeatletter
\def\@makefnmark{} 
\footnotetext{This work was supported by National Key R\&D Program of China (Grant No.2024YFB4708900). It was also supported in part by the Natural Science Foundation of
China under Grant 625B1026, 62225309, U24A20278, 62361166632.}

\footnotetext{$^1$ Shanghai Jiao Tong University, Shanghai 200240, China. }

\footnotetext{$^2$ University of California, Berkeley, Berkeley, CA 94720, USA. }

\footnotetext{$^*$ Equal contribution. $^{\dagger}$ Corresponding author.}
\def\@makefnmark{\hbox{\@textsuperscript{\normalfont\@thefnmark}}} 
\makeatother
}

\begin{abstract}

Urban scene reconstruction is critical for autonomous driving, enabling structured 3D representations for data synthesis and closed-loop testing. Supervised approaches rely on costly human annotations and lack scalability, while current self-supervised methods often confuse static and dynamic elements and fail to distinguish individual dynamic objects, limiting fine-grained editing. We propose \ourmethod, a novel dynamic instance-aware reconstruction method for label-free street scenes with 4D Gaussian Splatting. We first accurately identify dynamic instances by exploiting appearance–position inconsistency between warped rendering and actual observations. Guided by instance-level dynamic perception, we employ instance-aware 4D Gaussians as the unified volumetric representation, realizing dynamic-adaptive and instance-aware reconstruction. Furthermore, we introduce a reciprocal mechanism through which identity and dynamics reinforce each other, enhancing both integrity and consistency. Experiments on urban driving scenarios show that \ourmethod surpasses existing self-supervised baselines in reconstruction quality and instance-level editing, offering a concise yet powerful solution for urban scene modeling. Our code and models are available at: \url{https://github.com/IRMVLab/DIAL-GS}.

\end{abstract}
\section{INTRODUCTION}


Urban scene reconstruction has become a cornerstone technology in autonomous driving. By generating structured 3D representations of complex urban environments, it provides foundations for large-scale data synthesis, supporting both algorithm development and closed-loop testing in safety-critical scenarios~\cite{zhu2024scene}.

To address the challenges of road scenes, many existing methods adopt supervised learning, which relies on labor-intensive manual annotations to accurately capture the spatial and semantic information of dynamic objects~\cite{yan2024street,zhou2024drivinggaussian,chen2024omnire,khan2024autosplat,prosgnerf,zhou2024hugs,hwang2024vegs,hess2025splatad,ost2021neural}. However, manual labeling is costly, and supervised models are inherently limited to the scope of annotated datasets, hindering their scalability. These drawbacks have motivated increasing interest in self-supervised reconstruction~\cite{peng2024desiregs4dstreetgaussians,chen2023periodic,sun2025splatflow,mao2025unire}. 

\begin{figure}[t]
    \centering
    \includegraphics[width=1\linewidth]{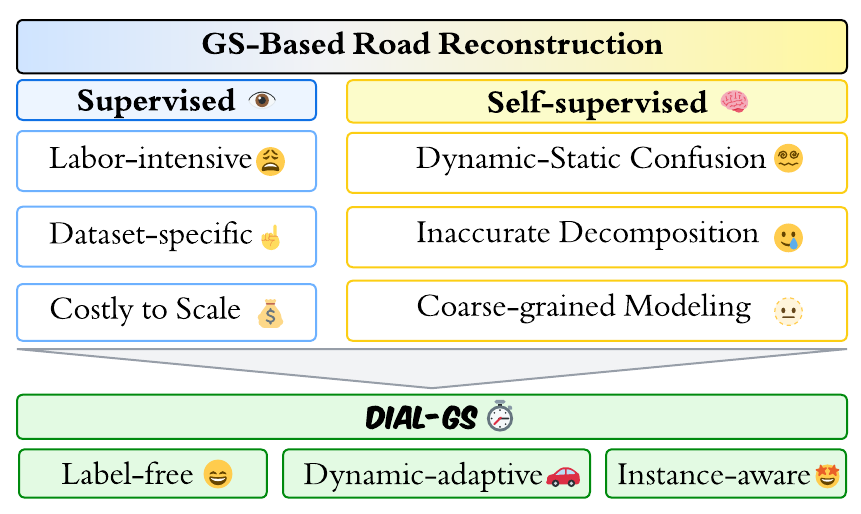}
    \caption{Motivation. DIAL-GS overcomes the limits of supervised and self-supervised methods with label-free, dynamic-adaptive and instance-aware reconstruction.}
    \label{fig1:motivation}
\end{figure}


Without explicit supervisory signals, self-supervised methods are prone to dynamic-static confusion: static objects may be incorrectly modeled as dynamic due to data noise, while slowly moving objects may be mistakenly treated as static. To address this issue, DIAL-GS introduces an inconsistency-driven approach
for precise instance-level dynamic perception. Specifically, when dynamic objects are forced to be represented by static Gaussians, the resulting static field merely records their instantaneous states in the past, which inevitably lag behind the current observations. This discrepancy manifests as the inconsistency between rendering and ground truth in both appearance and position, which DIAL-GS leverages as a reliable cue to distinguish dynamic instances from the static background.

Another challenge in self-supervised reconstruction lies in constructing a unified representation for the entire scene. Supervised paradigms often sidestep this difficulty by applying static Gaussians~\cite{kerbl20233d} to static background and employing time-varying Gaussians for dynamic objects~\cite{yan2024street, zhu2024scene, chen2024omnire}. In contrast, self-supervised frameworks cannot pre-identify elements as static or dynamic. This necessitates a representation that is simultaneously capable of preserving invariant spatial attributes of static background while modeling the spatiotemporal variations of dynamic objects. 

Beyond unified representation, a further difficulty of self-supervised reconstruction lies in enabling scene editing, which is fundamental for data generation. Unfortunately, current self-supervised approaches lack instance awareness, reducing the edition to coarse static–dynamic decomposition. Without the ability to distinguish between individual dynamic objects, these methods cannot support per-instance modeling or fine-grained editing, which severely limits their applicability.

DIAL-GS adopts instance-aware 4DGS to handle these limitations. By jointly encoding identity and dynamic attributes, DIAL-GS provides a unified framework where both static and dynamic scene components are consistently modeled and each Gaussian primitive is enriched with ID features. Furthermore, we propose a reciprocal ID–dynamics training strategy. We identify Gaussians belonging to the same instance via ID embeddings and enforce their dynamics consistency, while dynamic attributes are leveraged to select existing Gaussians and cluster their ID embeddings. In this way, the integrity of instance awareness and the consistency of dynamic modeling are jointly enhanced.

With these mechanisms, DIAL-GS realizes a dynamic-adaptive and instance-aware 4D reconstruction within the self-supervised regime. Our main contributions are summarized as follows:

\begin{enumerate}
    \item We introduce an intuitive and accurate instance-level dynamic perception algorithm by exploiting the appearance and position inconsistency caused by motion.
    \item We empower self-supervised reconstruction with instance awareness by proposing instance-aware 4DGS, and introduce a reciprocal mechanism in which instance awareness and dynamics mutually benefit.
    \item Extensive experiments demonstrate that \ourmethod surpasses prior methods in image reconstruction and novel view synthesis, while uniquely enabling instance-level editing -- a capability absent from existing self-supervised approaches.

\end{enumerate}

\section{RELATED WORK}

\subsection{NeRF-based Driving Reconstruction}
Numerous methods have investigated NeRF-based approaches for road-scene reconstruction. NSG~\cite{ost2021neural} models dynamic multi-object scenes with a scene graph, enabling instance-level view synthesis and 3D detection. Block-NeRF~\cite{tancik2022block} adopts block-wise representations with semantic masking and appearance codes to reconstruct large-scale scenes, while READ~\cite{li2023read} introduces a real-time rendering engine with $\omega$-net for photorealistic scene synthesis and editing. SUDS~\cite{turki2023suds} factorizes scenes into static, dynamic, and far-field radiance fields with hash-grid acceleration, supporting scene flow estimation and semantic manipulation. EmerNeRF~\cite{yang2023emernerf} advances self-supervised modeling by estimating flow-based correspondences and leveraging 2D foundation features for geometry and semantics.

Despite these advances, NeRF-based methods remain constrained by heavy computation, slow training and rendering, and limited scalability to large dynamic environments. Their reliance on dense sampling and implicit volumetric MLPs further hinders real-time applications and fine-grained editing~\cite{rabby2023beyondpixels}. By contrast, Gaussian Splatting achieves comparable visual quality with significantly faster performance and naturally accommodates instance-aware extensions.

\subsection{GS-based Driving Reconstruction}

3D Gaussian Splatting (3DGS)~\cite{kerbl20233d} has recently emerged as an efficient alternative to NeRFs, replacing implicit MLP-based volumetric fields with explicit Gaussian primitives. By rasterizing Gaussians, 3DGS enables fast rendering and, thanks to its explicit structure, naturally extends beyond view synthesis to tasks such as dynamic scene reconstruction, geometry editing, and physical simulation~\cite{wu2024recent}. These properties make 3DGS particularly well-suited for large-scale driving-scene reconstruction, where efficiency is essential.

Supervised methods rely on labor-annotated datasets to guide geometry, semantics, and dynamics, achieving highly accurate and structured reconstructions. DrivingGaussian~\cite{zhou2024drivinggaussian} combines incremental static reconstruction with a dynamic Gaussian graph for large-scale driving scenes. Street Gaussians~\cite{yan2024street} leverages a 4D spherical harmonics appearance model, tracked pose optimization, and point cloud initialization to improve rendering quality. AutoSplat~\cite{khan2024autosplat} introduces geometry-constrained background modeling, template-based foreground initialization, and temporally adaptive appearance modeling. OmniRe~\cite{chen2024omnire} builds a holistic scene graph that unifies static backgrounds, vehicles, SMPL-modeled humans~\cite{loper2023smpl}, and other non-rigid actors.
While effective, supervised approaches are labor-intensive, expensive to scale, and inherently constrained by the distribution of annotated datasets, limiting their generalization to unseen scenarios.

Self-supervised approaches remove the need for annotations by exploiting temporal consistency, geometric cues, and multi-view signals in driving data. $S^3$ Gaussians~\cite{huang2024textit} decomposes scenes with a spatio-temporal network that models Gaussian deformation through feature planes. PVG~\cite{chen2023periodic} introduces 4D Gaussians with periodic vibration and learnable lifespans, avoiding Hexplane-based deformation~\cite{cao2023hexplane}. DeSiRe-GS~\cite{peng2024desiregs4dstreetgaussians} enhances separation with a motion-mask extraction mechanism, 3D regularization, and temporal cross-view consistency.
Despite progress, these methods face two major limitations: (i) dynamic–static confusion, where noise, pose perturbations, or slow motion lead to misclassification of static versus dynamic objects, and (ii) lack of instance awareness, as they only coarsely separate static and dynamic components without distinguishing or editing individual objects. 

Our work directly addresses both issues by enabling accurate instance-level dynamic perception and introducing instance-aware 4D Gaussians.

\subsection{Semantic Scene Modeling with Gaussians}

Previous semantic scene modeling approaches are usually NeRF-based~\cite{zhi2021place,fu2022panoptic,mazur2022feature,peng2025q,siddiqui2023panoptic}. However, with the rise of 3D Gaussian Splatting (GS), increasing efforts focus on injecting 2D semantic knowledge into 3D-GS. Gaussian Grouping~\cite{ye2024gaussian} transfers SAM’s segmentation capability~\cite{kirillov2023segment}~\cite{ravi2024sam} into 3D, enabling zero-shot segmentation without 3D mask annotations. To resolve multi-granularity ambiguity, SAGA~\cite{cen2025segment} introduces a promptable 3D segmentation framework with a scale-gated mechanism and contrastive distillation. Semantic Gaussians~\cite{guo2024semantic} further projects diverse pre-trained 2D features into 3D Gaussians and enriches them with semantic attributes.

Building on these approaches, a natural direction for self-supervised road-scene reconstruction is to achieve instance awareness by embedding trajectory-tracked instance IDs into dynamic Gaussians. However, existing works are mostly tailored for static or small-scale scenes, limiting their applicability to highly dynamic driving environments. To bridge this gap, \ourmethod integrates ID-embedding with dynamic attributes and introduces a reciprocal training strategy that allow both to enhance each other.



\section{METHOD}

\begin{figure*}[t]
    \centering
    \includegraphics[width=1\linewidth]{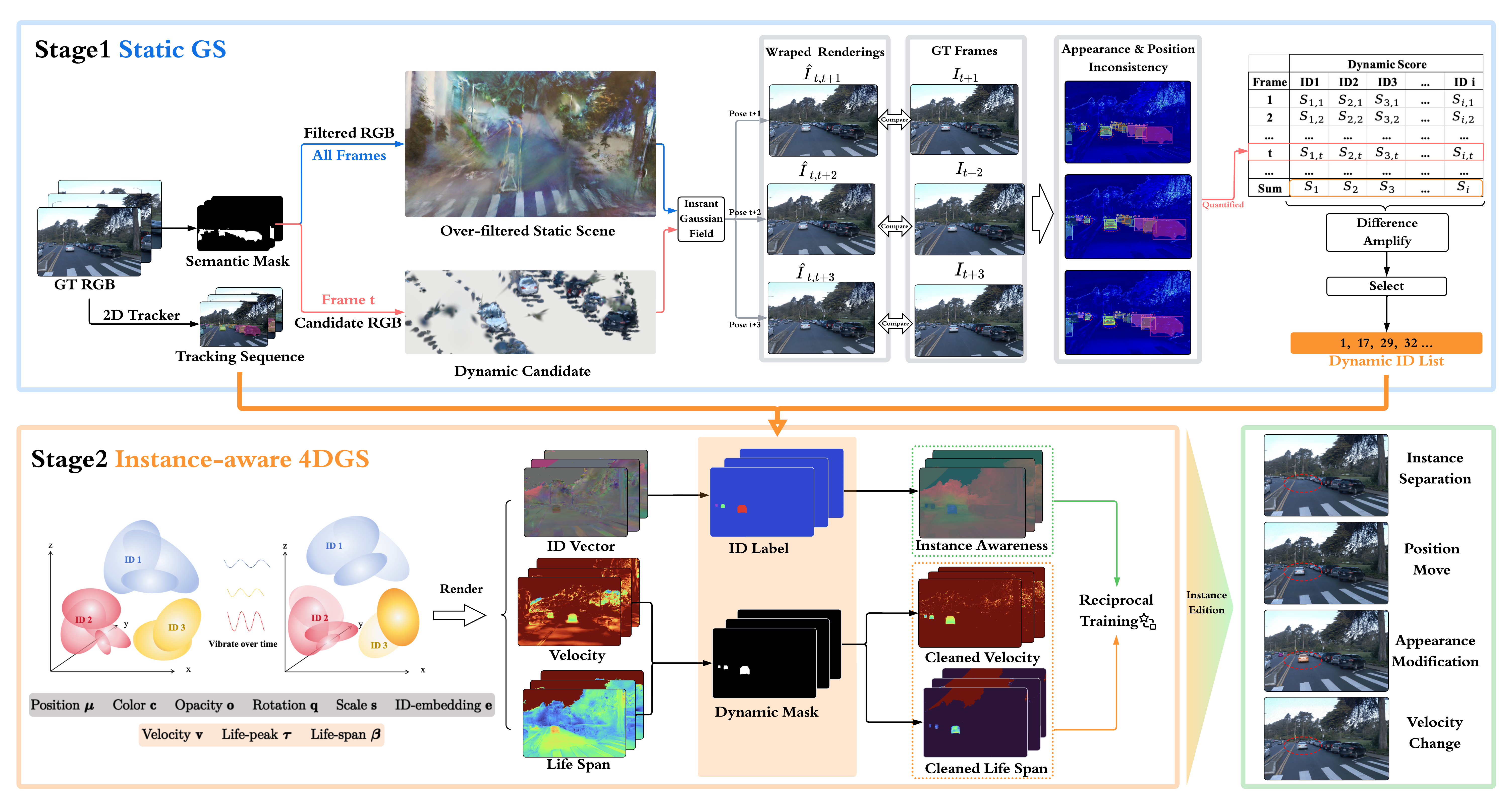}
    \caption{Method overview. (i)~Stage 1 conducts instance-level dynamic perception with static GS by exploiting inconsistency between warped renderings and ground-truth frames. Accumulated dynamic scores quantify inconsistency are used to obtain a dynamic ID list, according to which ID labels and dynamic masks are derived. (ii)~Stage 2 reconstructs the scene with instance-aware 4DGS as the unified representation. Guided by the ID labels and dynamic masks, it achieves instance awareness and refines dynamic attributes. Then it performs reciprocal training to enhance both instance awareness integrity and dynamics consistency. (iii)~With instance awareness, DIAL-GS further enables instance-level editing, a capability not supported by previous self-supervised approaches.}
    \label{fig:method}
\end{figure*}

We address the problem of road-scene reconstruction under a self-supervised setting. Given a sequence of temporally aligned multi-view observations—including RGB images, LiDAR point clouds, camera intrinsics, and ego-poses—our objective is to recover a scene representation that captures both static structures and dynamic objects without any 3D annotation.

As shown in Fig.~\ref{fig:method}, we design a two-stage pipeline. In Stage 1, we first reconstruct an over-filtered static scene with all frames and build dynamic candidate Gaussians for each frame. By combining them, we establish an instantaneous Gaussian field, where dynamic objects lag behind ground-truth observations and exhibit inconsistency in both appearance and position when rendered from new views. We quantify the inconsistency as a dynamic score for each instance and select dynamic IDs based on the aggregated scores. With the dynamic ID list, we then derive ID labels and dynamic masks from tracking sequence. In Stage 2, the scene is subsequently reconstructed with instance-aware 4D Gaussians as a unified representation. Guided by the labels and masks from Stage 1, we realize instance awareness and clean dynamic attributes, and further propose the reciprocal identity-dynamics training to reinforce the integrity of ID-embedding and consistency of dynamic attributes. With these designs, DIAL-GS enables self-supervised reconstruction to support instance-level editing.


\subsection{Instance-level Dynamic Perception}

When represented with static GS, only instantaneous states of dynamic objects can be captured. As time progresses, the previously recorded state of a dynamic object inevitably diverges from the actual observation, much like how a frozen image of a moving object always differs from its true current state. Building on this intuition, we propose Instance-level Dynamic Perception driven by inconsistency. 



\noindent \textbf{Instantaneous Gaussian Field Establishment.} We start by extracting the tracking sequence and semantic masks using BoT-SORT~\cite{aharon2022bot}. The semantic masks divide each frame into a filtered region (mainly static content, though some static objects may be excluded) and a candidate region (where potential dynamic objects remain). Using all filtered RGB frames together with all filtered LiDAR point clouds, we reconstruct an over-filtered static scene $S_{over}$ with static GS, which provides a temporally consistent background.

At time $t$, the point cloud is back-projected to recover candidates' spatial structure and, with RGB supervision,  dynamic candidate Gaussians $C_t$ is obtained. Combined with the static field $S_{over}$, they form the instantaneous Gaussian field, which is then warped to frames $t+1,\dots,t+k$ to simulate new-view observations:
\begin{equation}
   \hat{I}_{t,t+k}=\mathcal{F}_{t\to{t+k}}(C_t, S_{over}) ,
\end{equation} 
where $\mathcal{F}$ stands for the warp process.

\noindent \textbf{Dynamic Score and Instance Selection.} To measure the inconsistency of each candidate, we perform instance segmentation on $\hat{I}_{t,t+k}$ and establish ID correspondences with the ground-truth frame $I_{t+k}$. The appearance inconsistency of instance $i$ is defined as:
\begin{equation}
\mathcal{I}^{app}_{i,t} = \left \| I_{t+k} - \hat{I}_{t,t+k} \right \|\ \odot \frac{(\mathcal{M}_{i,t} \mid \hat{\mathcal{M}}_{i,t+k})}{\left \| (\mathcal{M}_{i,t} \mid \hat{\mathcal{M}}_{i,t+k}) \right \| },
\end{equation}
where $\mathcal{M}_{i,t}$ denotes the mask of instance $i$ at frame $t$, and $(\cdot \mid \cdot)$ represents the union of two masks.

Meanwhile, the position inconsistency is defined as
\begin{equation}
\begin{split}
\mathcal{I}^{pos}_{i,t} &= \frac{| c_{i,t} - \hat{c}_{i,t+k} |}{\sqrt{A_{i,t}}} + \frac{|A_{i,t} - \hat{A}_{i,t+k}|}{A_{i,t}}  +\frac{\sigma(E_{i,t,t+k})}{\sqrt{A_{i,t}}},
\end{split}
\end{equation}
where $c$ denotes the bounding box center, $A$ its area, $E_{i,t,t+k}$ the edge difference, and $\sigma$ the standard deviation.

By combining these two inconsistency measures, we define the dynamic score of instance $i$ at frame $t$ as
$S_{i,t} = \mathcal{I}^{app}_{i,t} + \mathcal{I}^{pos}_{i,t}.$ We compute $S_{i,t}$ from the first frame and accumulate them to obtain the final score:
$S_i = \frac{1}{n} \sum_t S_{i,t}$,
where $n$ denotes the total number of frames in which instance $i$ appears.  To enhance the separation between dynamic and static instances, we apply a cubic amplification and classify an instance as dynamic if $S_i^3 > \delta$. Finally, a dynamic ID list $\mathcal{D}=\{i|S_i^3 > \delta \}$ is obtained in stage one.

\subsection{Self-supervised Reconstruction with instance-aware 4DGS}

While existing self-supervised reconstruction methods have shown promise in capturing geometry and motion, they largely overlook instance awareness. Without distinguishing which Gaussian belongs to which object, these approaches often produce entangled representations that hinder reliable decomposition and fine-grained editing. This limitation motivates us to introduce instance-aware 4DGS as the unified representation.

\noindent \textbf{Instance-aware 4D Gaussian.} To consistently model both static and dynamic scene components and realize instance awareness, we embed ID vectors to PVG~\cite{chen2023periodic}. Thereby, the Gaussian of stage two is formulated with position $\boldsymbol{\mu}$, color $\mathbf{c}$, opacity $o$, rotation $\boldsymbol{q}$, scale $\boldsymbol{s}$, ID-embedding $\boldsymbol{e}$ as static attributes and velocity $\boldsymbol{v}$, life-peak $\tau$, life-span $\beta$ as dynamic attributes. The Gaussian vibrates around $\boldsymbol{\mu}$ and fades away according to $\tau$ and $ \beta$:

\begin{equation}
\tilde{\boldsymbol{\mu}}(t) = \boldsymbol{\mu} + \frac{l}{2\pi} \cdot \sin\left(\frac{2\pi(t - \tau)}{l}\right) \cdot \boldsymbol{v},
\end{equation}

\begin{equation}
\tilde{o}(t) = o \cdot exp({-\frac{1}{2}(t - \tau)^2 \beta^{-2})}.
\label{equation:dynamic_opacity}
\end{equation}

\noindent \textbf{Identity Loss.} We formulate the ID-embedding $\boldsymbol{e}$ as a static 8-bit vector inspired by \cite{ye2024gaussian} and render it with the original splatting pipeline.

\begin{equation}
    \mathbf{E}  = \sum_{i \in \mathcal{N}} \boldsymbol{e}_i \alpha_i \prod_{j=1}^{i-1} (1 - \alpha_j),
    \label{eq:contribution}
\end{equation}
where $\mathcal{N}$ is the total number of depth-sorted Gaussians, and $\boldsymbol{e}_i$ and $\alpha_i$ denote the ID embedding and density of the $i$-th Gaussian, respectively.

We then take a simple MLP $l$ followed by a softmax as the classifier: $\hat{I}^{id}=argmax(softmax(l(\mathbf{E})))$, where $\hat{I}^{id}$ stands for ID rendering. By ignoring instances not in $\mathcal{D}$, ID label $I^{id}$ can be derived from tracking sequence. The identity loss is then defined as the pixel-wise cross-entropy between the predicted ID renderings and the ID labels: 

\begin{equation}
    \mathcal{L}_{\mathrm{id}} = -\frac{1}{P} \sum_{p=1}^{P} \sum_{c=1}^{C} I^{id}_{p,c} \log\left( \hat{I}^{id}_{p,c} \right),
    \label{eq:id_loss}
\end{equation}
where $P=h\cdot w$ is pixel number of the rendering, $C$ is the max number of dynamic IDs, and $\hat{I}^{id}_{p,c} \in [0,1]$ is the predicted probability that pixel $p$ belongs to class $c$
and $I^{id}_{p,c} \in \{0,1\}$ is the one-hot ground-truth indicator.


\noindent \textbf{Dynamic Attributes Regularization.}
\label{sec:dynamic_reg}
PVG~\cite{chen2023periodic} tends to assign dynamics to static region inconsistent across time due to data noise, and fails to capture the dynamics of slowly moving objects. DeSiRe-GS~\cite{peng2024desiregs4dstreetgaussians} attempts to alleviate this issue by introducing a motion mask, yet it still inherits similar confusion since the mask itself is also learned in a fully self-supervised manner. In contrast, DIAL-GS leverages 2D trackers to directly generate instance mask and derive dynamic mask $\mathcal{M}$ by selecting IDs in $\mathcal{D}$. We then regulate dynamics with precise and instance-level mask:
\begin{equation}
    \mathcal{L}_{\bar{v}} = \frac{1}{|\mathcal{\bar{M}}|}I^{\bar{v}} \odot \mathcal{\bar{M}},
\end{equation}

\begin{equation}
     \mathcal{L}_{\beta} = -\frac{1}{|\mathcal{\bar{M}}|}I^{\beta} \odot \mathcal{\bar{M}},
\end{equation}
where $\bar{\boldsymbol{v}} = \boldsymbol{v} \cdot exp(-\frac{\beta}{2l})$ represents the instant velocity, and $I^{\bar{v}}$, $I^{\beta}$ represent the rendering of $\bar{\boldsymbol{v}}$ and $\beta$ respectively.

\begin{figure}[tb]
    \centering
    \includegraphics[width=1\linewidth]{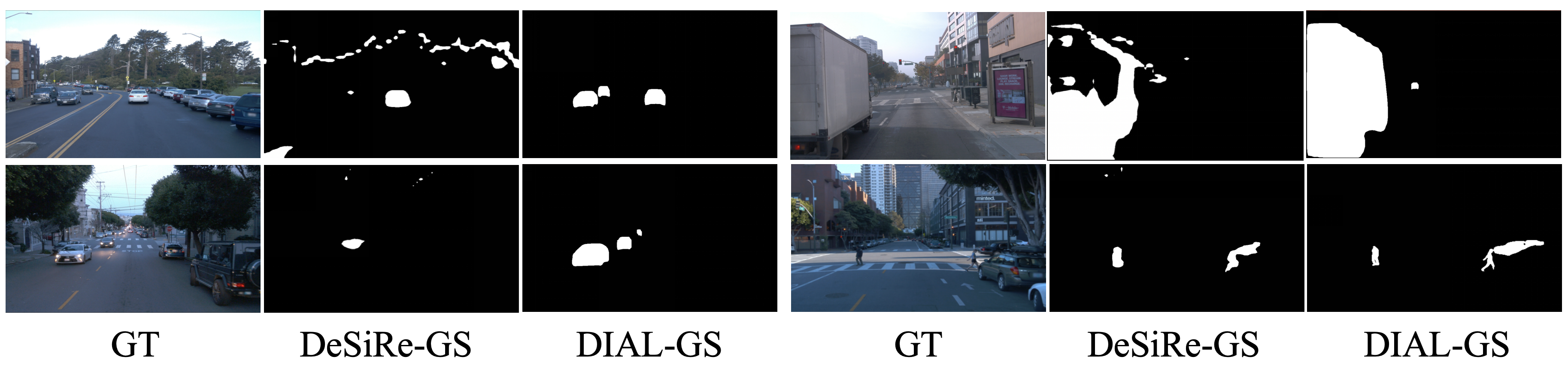}
    \caption{Dynamic Mask Comparison. DeSiRe-GS 
    misclassifies static region and incompletely captures dynamic parts. DIAL-GS obtains accurate and sharp dynamic masks instead.}
    \label{fig:mask_compare}
\end{figure}

\begin{table*}[t]
    \centering
    \caption{Comparison of methods on the Waymo Open Dataset and KITTI Dataset.}
    \label{tab:all_results}
    \setlength{\tabcolsep}{2pt}
    \resizebox{\textwidth}{!}{%
    \begin{tabular}{l ccc ccc ccc ccc }
        \toprule
       \multirow{3}{*}{\textbf{Method}}  & \multicolumn{6}{c}{\textbf{Waymo Open Dataset}} & \multicolumn{6}{c}{\textbf{KITTI}} \\
    \cmidrule(lr){2-7} \cmidrule(lr){8-13} 
         & \multicolumn{3}{c}{{Image reconstruction}} & \multicolumn{3}{c}{{Novel view synthesis}} & \multicolumn{3}{c}{{Image reconstruction}} & \multicolumn{3}{c}{{Novel view synthesis}}  \\
        \cmidrule(lr){2-4} \cmidrule(lr){5-7}  \cmidrule(lr){8-10} \cmidrule(lr){11-13} 
         & PSNR $\uparrow$ & SSIM $\uparrow$ & LPIPS $\downarrow$ & PSNR $\uparrow$ & SSIM $\uparrow$ & LPIPS $\downarrow$ & PSNR $\uparrow$ & SSIM $\uparrow$ & LPIPS $\downarrow$ & PSNR $\uparrow$ & SSIM $\uparrow$ & LPIPS $\downarrow$ \\
        \midrule
        S-NeRF~\cite{xie2023s}  & 19.67 & 0.528 & 0.387 & 19.22 & 0.515 & 0.400  & 19.23 & 0.664 & 0.193 & 18.71 & 0.606 & 0.352  \\
        StreetSurf~\cite{guo2023streetsurf}  & 26.70 & 0.846 & 0.3717 & 23.78 & 0.822 & 0.401 & 24.14 & 0.819 & 0.257 & 22.48 & 0.763 & 0.304  \\
        3DGS~\cite{kerbl20233d}    & 27.99 & 0.866 & 0.293 & 25.08 & 0.822 & 0.319 & 21.02 & 0.811 & 0.202 & 19.54 & 0.776 & 0.224  \\
        NSG~\cite{ost2021neural}   & 24.08 & 0.656 & 0.441 & 21.01 & 0.571 & 0.487 & 19.19  & 0.683 & 0.189 & 17.78 & 0.645 & 0.312  \\
        Mars~\cite{wu2023mars}   & 21.81 & 0.681 & 0.430 & 20.69 & 0.636 & 0.453  & 27.96 & 0.900 & 0.185 & 24.31 & 0.845 & 0.160  \\
        SUDS~\cite{turki2023suds}   & 28.83 & 0.805 & 0.317 & 25.36 & 0.783 & 0.384  & 28.83 & 0.917 & 0.147 & 26.07 & 0.797 & 0.131  \\
        EmerNeRF~\cite{yang2023emernerf}  & 28.11 & 0.786 & 0.373 & 25.92 & 0.763 & 0.384  & 26.95 & 0.828 & 0.218 & 25.24 & 0.801 & 0.237  \\
        PVG~\cite{chen2023periodic}  & {32.46} & {0.910} & {0.229} & {28.11} & {0.849} & {0.279}  & {32.83} & {0.937} & {0.070} & {27.43} & {0.896} & {0.114}  \\ 
        
        DeSiRe-GS~\cite{peng2024desiregs4dstreetgaussians}  & \underline{33.61} & \underline{0.919} & \underline{0.204} & \underline{29.75} & \underline{0.878} & \underline{0.213} & \underline{33.94} & \underline{0.949} & \underline{0.040} & \textbf{ 28.87 }& \underline{0.901} & \underline{0.106} \\
        
        \midrule
        \textbf{Ours} &  \textbf{36.88} & \textbf{0.948} &\textbf{ 0.113} & \textbf{30.14} & \textbf{0.880} & \textbf{0.183} & \textbf{34.23} & \textbf{0.954} & \textbf{0.039} & \underline{28.34} & \textbf{0.911} & \textbf{0.084} \\
        \bottomrule
    \end{tabular}
    }
\end{table*}

\begin{figure*}[t]
    \centering
    \includegraphics[width=1\linewidth]{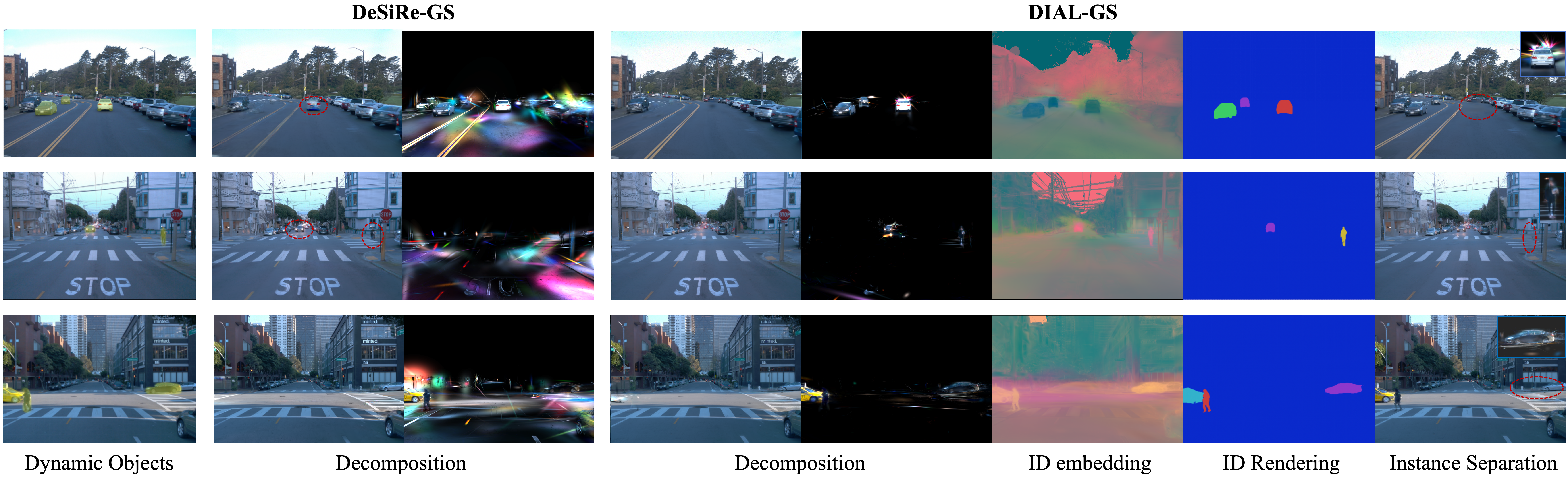}
    \caption{Qualitative results. Decomposition with DeSiRe-GS~\cite{peng2024desiregs4dstreetgaussians} suffers from severe misclassification, whereas DIAL-GS achieves accurate decomposition and clear instance separation.}
    \label{fig:decomposition}
\end{figure*}

\subsection{Reciprocal Identity-Dynamics Training}


Another limitation of existing self-supervised methods is the inconsistency of dynamic attributes among Gaussians belonging to the same object. For example, different Gaussians of a single car may exhibit significantly different instant velocities or life-spans. Moreover, relying solely on $\mathcal{L}_{id}$ often results in incomplete instance awareness, as it only provides 2D supervision. To address this, DIAL-GS introduces a reciprocal training scheme, enabling both more complete ID-embedding and more coherent dynamic attributes.

\noindent \textbf{3D Identity Loss.} Due to the mechanism of PVG~\cite{chen2023periodic}, Gaussians fade out by decreasing opacity while retaining their static attributes such as position. Therefore, clustering all ID embeddings would incur prohibitive computation, so we leverage dynamic attributes to filter Gaussians that actually exist in the current frame and perform ID clustering only on them.

We select existing Gaussians of frame $t$ as $\mathcal{G}^{exist}_t=\{j|\tilde{o}(t)>\epsilon\}$. For each Gaussian $j \in \mathcal{G}^{exist}_t$, we extract its predicted distribution as $P_j = softmax(l(\mathbf{e}_j))$ and predicted ID as $ID_j=argmax(P_j)$. We collect dynamic Gaussians as $\mathcal{G}^{dyn}_t=\{j|ID_j\in\mathcal{D},j\in\mathcal{G}^{exist}_t \}$, and static Gaussians as: $\mathcal{G}^{static}_t = \mathcal{G}^{exist}_t - \mathcal{G}^{dyn}_t$. For each Gaussian in $\mathcal{G}^{dyn}_t$, we then search for its $K$ nearest neighbors in $\mathcal{G}^{static}_t$ and obtain their predicted distributions $Q_k$. Finally, we adopt the KL divergence as the loss function to encourage nearby static Gaussians to align with the ID-embedding of dynamic Gaussians:
\begin{equation}
\mathcal{L}_{3d} = \frac{1}{K \cdot |\mathcal{G}^{dyn}_t|} \sum_{j=1}^{|\mathcal{G}^{dyn}_t|} \sum_{k=1}^K D_{KL}(P_j||Q_k).
\end{equation}

Note that $\mathcal{L}_{3d}$ is activated in the later stage of training, when both the dynamic attributes and the ID-embedding have nearly converged. Such a scheduling strategy maximizes the reliability of $\mathcal{G}^{exist}_t$ while exploiting the stable ID-embedding optimized by $\mathcal{L}_{id}$. By enforcing 3D clustering, instance awareness is no longer restricted to 2D alignment but instead achieves a holistic embedding in the 3D space.


\noindent \textbf{Dynamic Consistency Loss.} After the ID-embedding has stabilized through $\mathcal{L}_{id}$ and $\mathcal{L}_{3d}$, we leverage reliable instance awareness to enforce consistency of dynamic attributes. Similar to the computation of $\mathcal{L}_{3d}$, we first extract $\mathcal{G}^{exist}_t$ and their predicted IDs. For each instance in $\mathcal{D}$, we gather its Gaussians by $\mathcal{G}^{i}_t = \{ j \,|\, ID_j=i, i \in \mathcal{D} \}$. For every Gaussian in $\mathcal{G}^{i}_t$, we then search $K$ nearest neighbors within $\mathcal{G}^{i}_t$ and obtain their instant velocities and life-spans. The consistency losses are defined as follows:
\begin{equation}
    \mathcal{L}_{\text{mag}} = \frac{1}{K \cdot |\mathcal{G}^{i}_t|} \sum_{j=1}^{|\mathcal{G}^{i}_t|} \sum_{k=1}^K \frac{\| \bar{\boldsymbol{v}}_j - \bar{\boldsymbol{v}}_k \|_2}{{\|\bar{\boldsymbol{v}}_{mean}}\|_2 },
\end{equation}
\begin{equation}
    \mathcal{L}_{\text{dir}} = \frac{1}{2 K \cdot|\mathcal{G}^{i}_t|} \sum_{j=1}^{|\mathcal{G}^{i}_t|} \sum_{k=1}^K \left(1 - \frac{\bar{\boldsymbol{v}}_j \cdot \bar{\boldsymbol{v}}_k}{\|\bar{\boldsymbol{v}}_j\|_2 \cdot \|\bar{\boldsymbol{v}}_k\|_2 }\right),
\end{equation}
\begin{equation}
    \mathcal{L}_{\text{beta}} = \frac{1}{K \cdot |\mathcal{G}^{i}_t|} \sum_{j=1}^{|\mathcal{G}^{i}_t|} \sum_{k=1}^K \frac{|\beta_j - \beta_k|}{\bar{\beta}},
\end{equation}
\begin{equation}
    \mathcal{L}_{\text{consist}} = \lambda_{\text{mag}} \cdot \mathcal{L}_{\text{mag}} + \lambda_{\text{dir}} \cdot \mathcal{L}_{\text{dir}} + (1 - \lambda_{\text{mag}} - \lambda_{\text{dir}}) \cdot  \mathcal{L}_{\text{beta}}.
\end{equation}

The reinforcement of consistency encourages the 4DGS of the same dynamic object to form a coherent representation, thereby reducing artifacts in novel view synthesis (NVS) and allowing dynamic attributes to serve as reliable auxiliary cues besides ID-embedding for decomposition.


\subsection{Optimization}

In the first stage, all Gaussians are treated as static and the training loss consists: 

\begin{equation}
    \mathcal{L}_I = (1 - \lambda_{ssim}) \| I - \tilde{I} \| + \lambda_{ssim} \mathrm{SSIM}(I, \tilde{I}),
\end{equation}

\begin{equation}
    \mathcal{L}_D = \| I^{D} - D_{gt} \|,
\end{equation}

\begin{equation}
     \mathcal{L}_o = -\frac{1}{hw} \sum O \cdot \log O - \frac{1}{hw} \sum \mathcal{M}_{sky} \cdot \log(1 - O).
\end{equation}

The whole loss function of stage one is:
\begin{equation}
    \mathcal{L}_\text{stage1} = \lambda_I\mathcal{L}_I + \lambda_D\mathcal{L}_D+\lambda_o\mathcal{L}_O.
\end{equation}

In the second stage, we gradually introduce the proposed losses and  the overall objective of stage 2 is formulated as:

\begin{equation}
\begin{split}
\mathcal{L}_{\text{stage2}} = \lambda_I\mathcal{L}_{I} + \lambda_D\mathcal{L}_{D} + \lambda_o\mathcal{L}_O + \lambda_{\bar{v}}\mathcal{L}_{\bar{v}} + \lambda_{\beta}\mathcal{L}_{\beta} + \\ \lambda_{id}\mathcal{L}_{id} + \lambda_{3d}\mathcal{L}_{3d} + \lambda_{\text{consist}}\mathcal{L}_{\text{consist}}.
\end{split}
\end{equation}

\subsection{Instance-level Scene Edition}
Without instance awareness, former self-supervised works can only perform coarse decomposition. On the contrary, DIAL-GS empowers self-supervised reconstruction with the ability to edit specific instances.

We jointly consider the $\mathbf{e}$, $\beta$, $\bar{\boldsymbol{v}}$ and $\tilde{o}$ to select the Gaussians belonging to instance $i$ at frame $t$. By modifying $\tilde{\boldsymbol{\mu}}(t)$ and color $\mathbf{c}$, we change its position and appearance. We also use $\hat{\boldsymbol{\mu}}= \tilde{\boldsymbol{\mu}}(t) +\Delta t \cdot \bar{\boldsymbol{v}}$ to change instance's velocity while keeping its original trajectory.

\begin{figure}[t]
    \centering
    \includegraphics[width=0.8\linewidth]{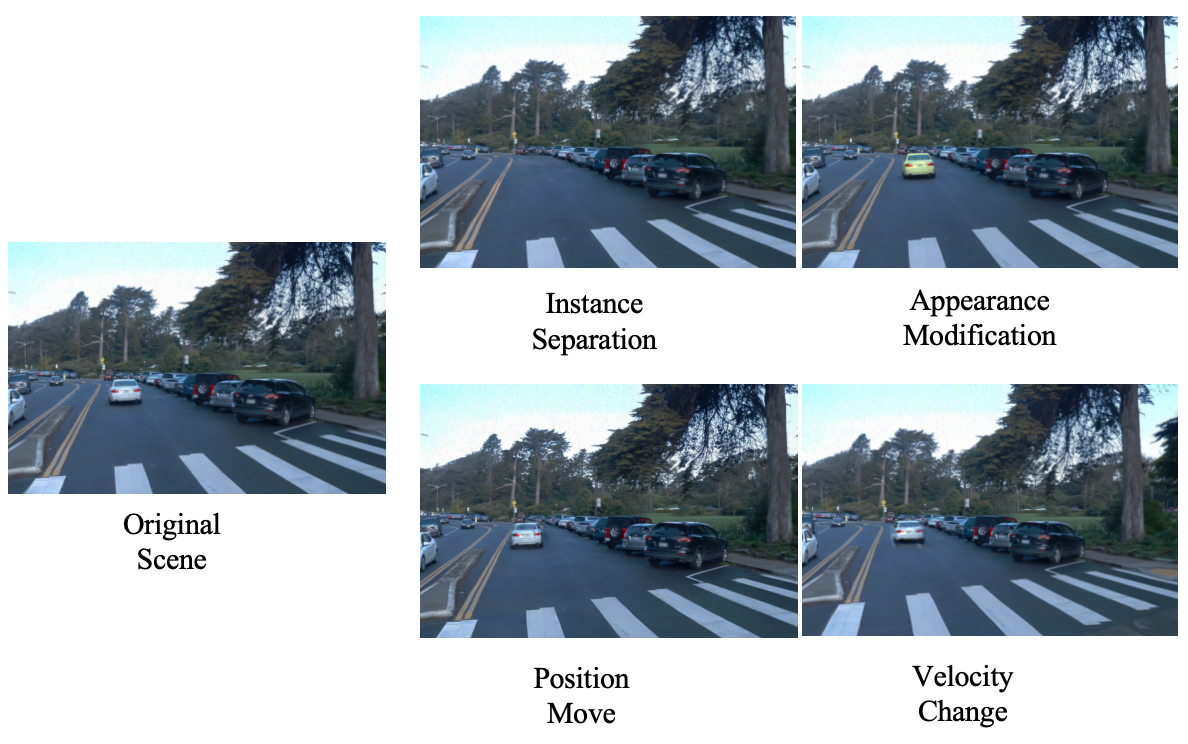}
    \caption{Instance Edition. By realizing instance awareness, DIAL-GS supports instance edition within the self-supervised regime.}
    \label{fig:placeholder}
\end{figure}

\section{Experiment}
\label{sec:experiment}

\subsection{Experimental Setting}
\label{sec:exp_set}
\noindent \textbf{Dataset.} We follow the experimental setup of PVG~\cite{chen2023periodic} and DeSiRe-GS~\cite{peng2024desiregs4dstreetgaussians}, focusing on highly dynamic scenarios to enable comprehensive baseline comparisons.

\noindent \textbf{Evaluation Metrics.} We adopt PSNR, SSIM and LPIPS as metrics for the evaluation of image rendering quality.

\noindent \textbf{Implementation Details.} We trained our model on one NVIDIA A100 Tensor Core GPU. In the first stage, we use BoT-SORT~\cite{Brostrom_BoxMOT_pluggable_SOTA,aharon2022bot,matterport_maskrcnn_2017} as the 2D tracker. We train $S_{over}$ for 30,000 iterations and each $C_t$ for 400 iterations. For efficiency and stability, we warp 2 frames to check inconsistency. We take $\delta$ as $1e-3$ to select dynamic IDs and $\epsilon$ as $5e-4$ to select existing Gaussians. During the 55,000 iterations of stage 2, we gradually introduce $\mathcal{L}_{2d}$, $\mathcal{L}_{\bar{v}}$, $\mathcal{L}_{\beta}$, $\mathcal{L}_{3d}$ and $\mathcal{L}_{consist}$ in sequence. We set K to 5 for KNN search. Weights mentioned are $\lambda_{ssim}=1$, $\lambda_{I}=1$, $\lambda_{D}=1$, $\lambda_{o}=0.05$, $\lambda_{\bar{v}}=0.01$, $\lambda_{\beta}=0.001$, $\lambda_{id}=0.1$, $\lambda_{3d}=1.5$, $\lambda_{\text{consist}}=0.01$, $\lambda_{\text{mag}}=0.4$ and $\lambda_{\text{dir}}=0.2$.

\subsection{Experimental Results}
\label{sec:exp_res}

We report quantitative results on Waymo Open Dataset~\cite{waymo_2020_cvpr} and KITTI~\cite{kitti_2012_cvpr} in Tab.~\ref{tab:all_results} in both image reconstruction and novel view synthesis. Ours achieves the best performance in most scenarios and demonstrates significant improvement especially in Waymo. In Fig.~\ref{fig:decomposition}, we provide qualitative comparison against DeSiRe-GS~\cite{peng2024desiregs4dstreetgaussians}. DeSiRe-GS relies on staticness coefficient for decomposition. It can be observed that DeSiRe-GS~\cite{peng2024desiregs4dstreetgaussians} suffers severe dynamic misclassification: artifacts of dynamic objects remain in the static part, small or slow-moving dynamic objects are mistakenly treated as static, and the dynamic part includes clearly static elements such as road surfaces and parked cars. By contrast, DIAL-GS employs identity as the primary criterion, with the staticness coefficient as auxiliary support, thereby achieving precise decomposition and enabling instance-level separation — a capability that prior self-supervised methods could not realize.

\subsection{Ablation Study}
\label{sec:exp_abl}
We conduct various ablation studies to verify the effectiveness of the instance-level dynamic perception and loss functions.

\begin{figure}[b]
    \centering
    \includegraphics[width=0.7\linewidth]{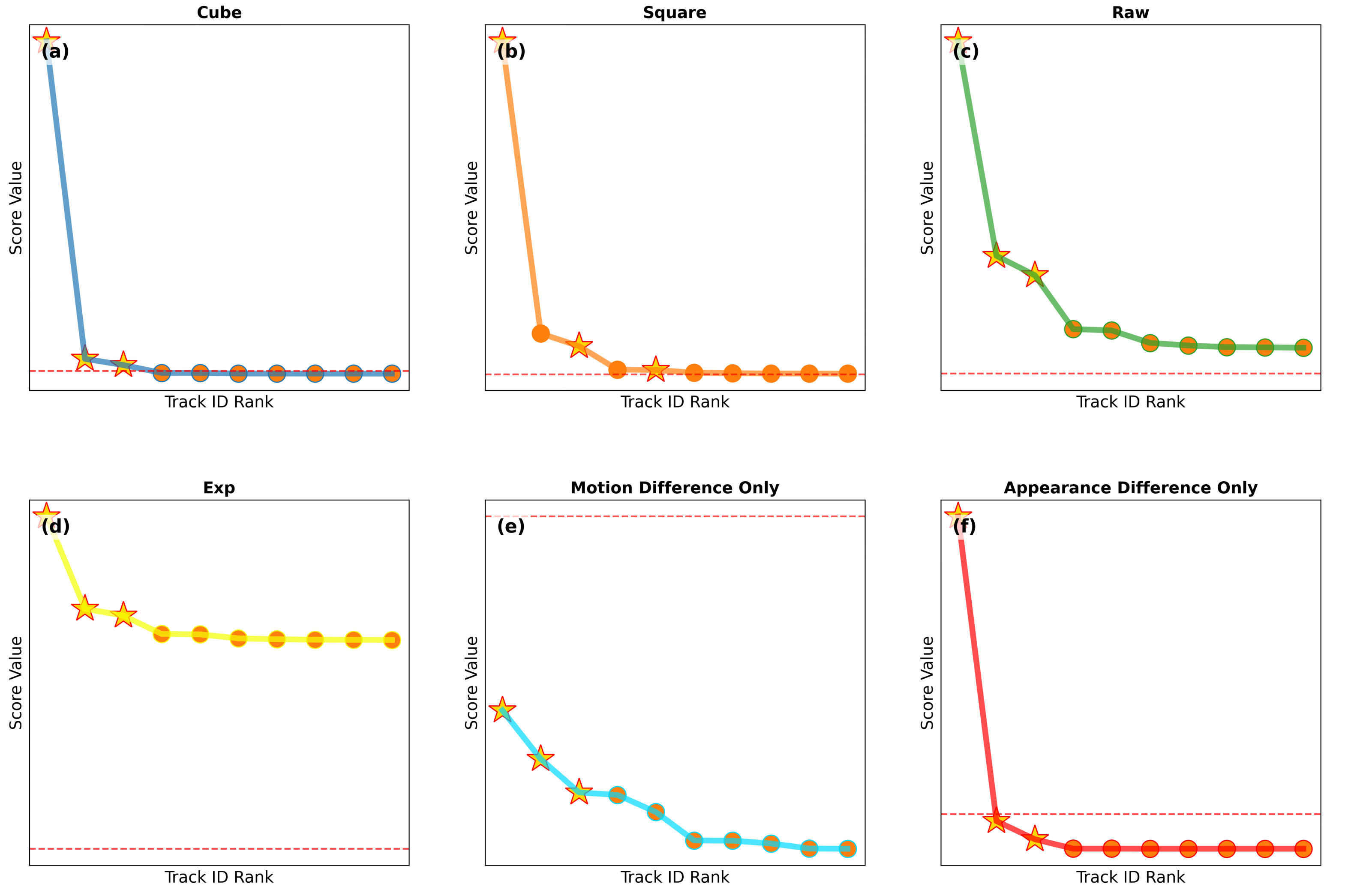}
    \caption{Ablation studies on dynamic score. Axes denote IDs and their scores. Stars represent dynamic instances, while circles represent static ones. Red line represents the threshold.}
    \label{fig:score_ablation}
\end{figure}

\begin{figure}[t]
    \centering
    \includegraphics[width=0.8\linewidth]{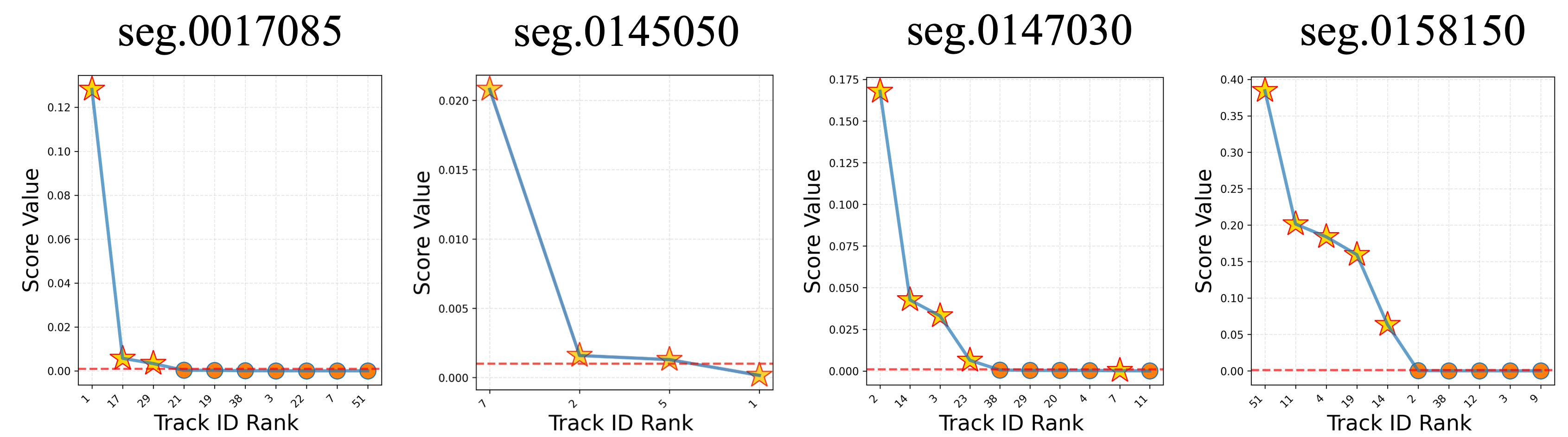}
    \caption{Dynamic ID selection in different scenes. Axes and icons follow the same convention as in Fig.~\ref{fig:score_ablation}.}
    \label{fig:diff_scene}
\end{figure}

\noindent \textbf{Dynamic score}. In stage one, we test different methods to select dynamic IDs within the score ranking list. As shown in Fig.~\ref{fig:score_ablation}, subfigures (a)–(d) present the dynamic score ranking using $S_i^3$, $S_i^2$, $S_i$, and $\exp(S_i)$, respectively. Subfigures (e) and (f) further analyze $S_i^3$ when computed from only one type of inconsistency: $S_{i,t}^{(e)}=\mathcal{I}^{pos}_{i,t}$ and $S_{i,t}^{(f)}=\mathcal{I}^{app}_{i,t}$. The results demonstrate that stable separation between dynamic and static IDs is realized by jointly considering appearance and position inconsistency with the cubic transform. We evaluate the selection procedure across different scenes. As shown in Fig.~\ref{fig:diff_scene}, the cubic transformation suppresses static-instance scores toward zero while preserving relatively high scores for dynamic instances, enabling stable thresholding for dynamic ID separation. The few remaining misclassifications mainly arise from distant or small objects with limited observations.

\begin{table}[t]
\centering
\caption{Ablation Study with Loss Configurations.}
\label{tab:ablation}
\setlength{\tabcolsep}{4pt}
\begin{tabular}{c|ccccc|ccc}
\hline
\textbf{Exp.} & $\mathcal{L}_{\bar{v}}$ & $\mathcal{L}_{\beta}$ & $\mathcal{L}_{id}$ & $\mathcal{L}_{3d}$ & $\mathcal{L}_{\text{consist}}$ & PSNR$\uparrow$ & SSIM$\uparrow$ & LPIPS$\downarrow$ \\
\hline
(a) & -- & \checkmark & \checkmark & \checkmark & \checkmark & 29.9262 & 0.8792 & 0.1817 \\
(b) & \checkmark & -- & \checkmark & \checkmark & \checkmark & 29.8243 & 0.8674 & 0.1989 \\
(c) & \checkmark & \checkmark & -- & -- & -- & 29.8331 & 0.8788 & 0.1833 \\
(d) & \checkmark & \checkmark & \checkmark & -- & \checkmark & 29.8405 & 0.8788 & 0.1823 \\
(e) & \checkmark & \checkmark & \checkmark & \checkmark & -- & 29.9032 & 0.8792 & \textbf{0.1813} \\
(f) & \checkmark & \checkmark & \checkmark & \checkmark & \checkmark & \textbf{30.1368} & \textbf{0.8803} & 0.1826 \\
\hline
\end{tabular}
\end{table}

\begin{figure}[tb]
    \centering
    \includegraphics[width=1\linewidth]{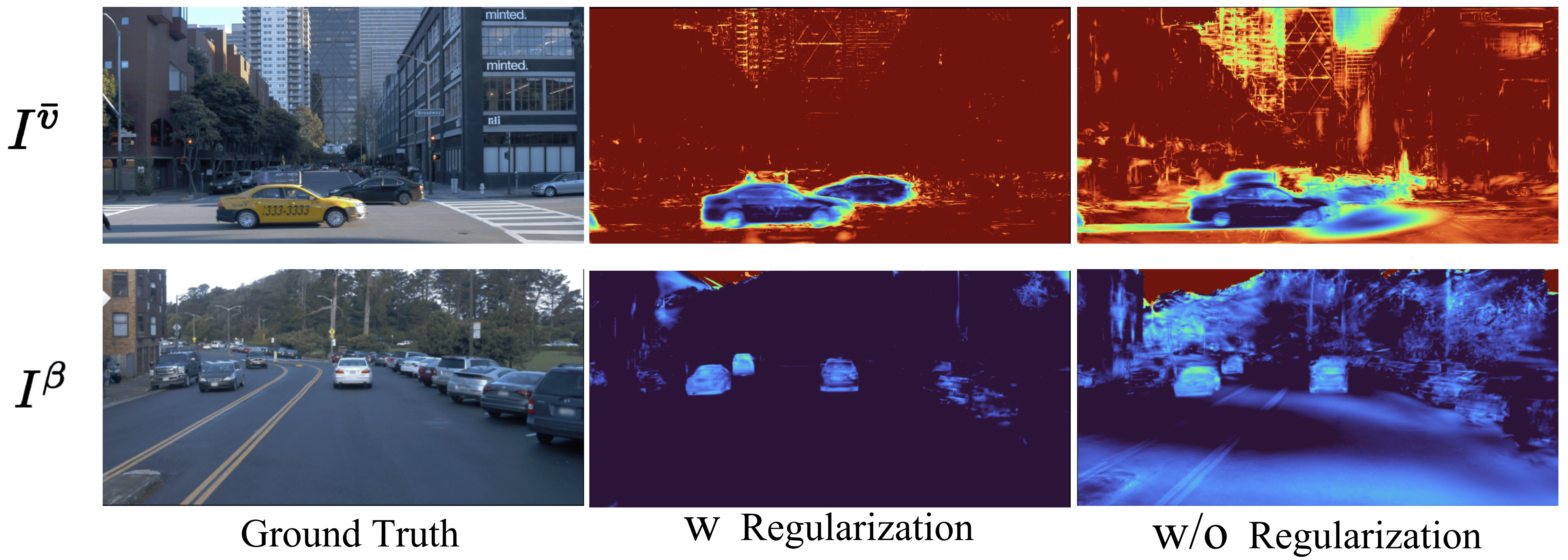}
    \caption{Effect of Dynamic Attribute Regularization. The regularization ensures clean static regions while constraining velocity and life-span to avoid excessive growth or shrinkage.}
    \label{fig:reg}
\end{figure}

\noindent \textbf{Dynamic Attribute Regularization.} We ablate $\mathcal{L}{\bar{v}}$ and $\mathcal{L}_{\beta}$ to examine the role of velocity and life-span regularization. As shown in Tab.~\ref{tab:ablation}(a)(b), both losses contribute to improving novel-view rendering quality. Moreover, Fig.~\ref{fig:reg} illustrates that removing the regularization leads to inaccurate dynamic attributes in distant static regions and velocity grows excessively and life-span shrinks, ultimately causing overfitting to training views.

\noindent \textbf{ID-embedding.} The result in Tab.~\ref{tab:ablation}(c) shows that ID-embedding not only enables instance awareness but also improves the reconstruction quality.

\begin{figure}[t]
    \centering
    \includegraphics[width=1\linewidth]{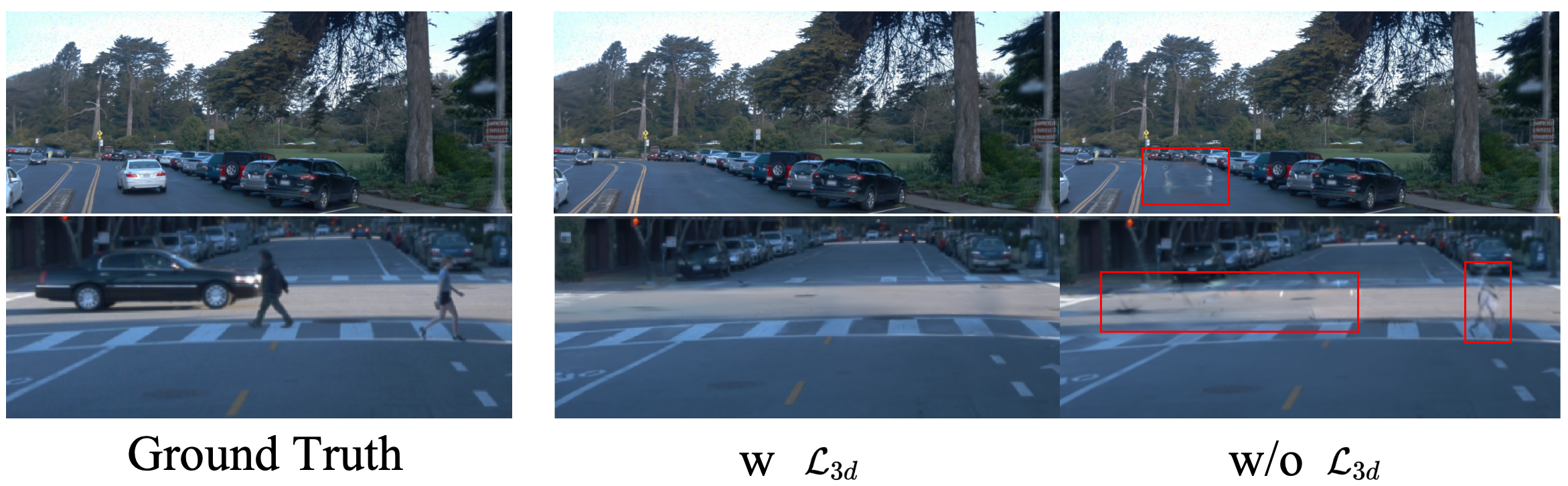}
    \caption{Effect of $\mathcal{L}_{3d}$. With $\mathcal{L}_{3d}$, the ID-embedding is more complete and the decomposition is more clean.}
    \label{fig:3d_loss}
\end{figure}

\begin{figure}[t]
    \centering
    \includegraphics[width=1\linewidth]{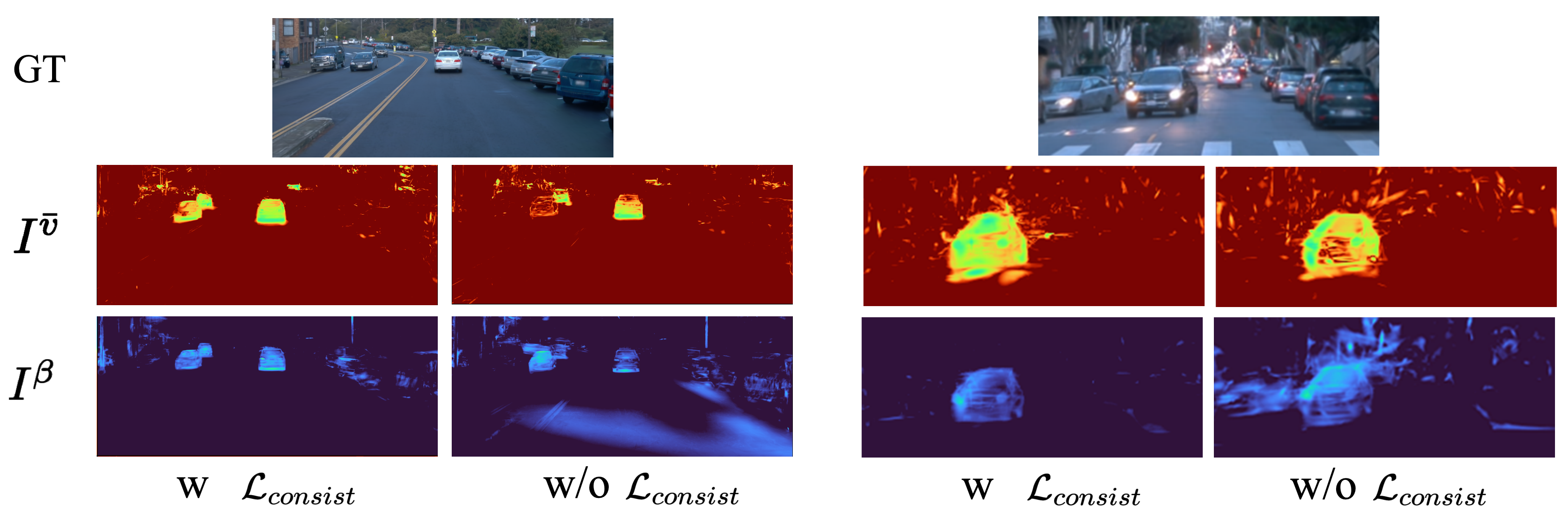}
    \caption{Effect of $\mathcal{L}_{\text{consist}}$. The dynamic attributes are more consistent and reasonable with the guide of $\mathcal{L}_{\text{consist}}$.}
    \label{fig:consist}
\end{figure}

\noindent \textbf{Reciprocal Identity–Dynamics Training.} From Tab.~\ref{tab:ablation}(d)(e), we observe that the NVS quality drops when either $\mathcal{L}_{3d}$ or $\mathcal{L}_{\text{consist}}$ is removed. Fig.~\ref{fig:3d_loss} shows that removing $\mathcal{L}_{3d}$ leads to incomplete ID embeddings, resulting in residuals along vehicle boundaries (first row) and artifacts within dynamic objects (second row). Fig.~\ref{fig:consist} shows that $\mathcal{L}_{\text{consist}}$ enhances the consistency of dynamic attributes within the same instance. As a result, the motion modeling adheres more closely to realistic physical behavior, which in turn improves both novel view synthesis and decomposition.

\subsection{Discussion}

While DIAL-GS achieves strong instance-aware reconstruction, it relies on external 2D tracker, which may introduce occasional errors. We nonetheless adopt this design because fully self-supervised dynamic perception remains unreliable for complex road scenes as shown by DeSiRe-GS~\cite{peng2024desiregs4dstreetgaussians} and discussed in Sec.~\ref{sec:dynamic_reg}. Furthermore, generic self-supervised features~\cite{yue2024improving} adapted by DeSiRe-GS~\cite{peng2024desiregs4dstreetgaussians} are not tailored to driving environments and lack awareness of object-motion patterns. In contrast, modern 2D trackers have already encoded rich semantic and physical priors specific to traffic scenes, producing more precise instance masks than those obtained from purely self-supervised approaches, making them a pragmatic and effective choice for our framework.

\section{Conclusion}
    In this paper, we present DIAL-GS, a novel self-supervised framework with dynamic instance awareness. DIAL-GS achieves instance-level dynamic perception by leveraging inconsistency caused by motion. By proposing instance-aware 4DGS, DIAL-GS jointly encodes identity and dynamic attributes and further enables them to benefit each other in reciprocal identity–dynamics training strategy. Extensive experiments validate its effectiveness, demonstrating that DIAL-GS advances self-supervised reconstruction for real-world autonomous driving scenarios.

\clearpage

\bibliographystyle{./IEEEtran}
\bibliography{./IEEEexample}

\end{document}